\documentclass[runningheads]{llncs}
\usepackage[T1]{fontenc}
\usepackage{amsmath}
\usepackage{cite}
\usepackage{todonotes}
\usepackage{marvosym}
\usepackage{hyperref}
\hypersetup{
    colorlinks=true,        
    linkcolor=magenta,      
    citecolor=magenta,      
    filecolor=magenta,      
    urlcolor=magenta        
}
\usepackage{graphicx}
\usepackage{subfig}
\usepackage{array}
\usepackage{booktabs}

\usepackage{color}

\begin{document}
\title{Beyond Known Reality: Exploiting Counterfactual Explanations for Medical Research}

\author{
Toygar Tanyel\inst{1}\textsuperscript{(\Letter)} \and
Serkan Ayvaz\inst{2}\textsuperscript{(\Letter)} \and
Bilgin Keserci\inst{3}
}

\institute{
Biomedical Engineering Graduate Program, Istanbul Technical University \and 
Centre for Industrial Software, Maersk Mc-kinney Moeller Institute, University of Southern Denmark\and
Department of Biomedical Engineering, Yildiz Technical University \\
(\Letter) \email{tanyel23@itu.edu.tr}  \: (\Letter) \email{seay@mmmi.sdu.dk} 
}

\authorrunning{Tanyel et al.}
%
%
\titlerunning{Exploiting Counterfactual Explanations for Medical Research}

\maketitle              

\begin{abstract}
The field of explainability in artificial intelligence (AI) has witnessed a growing number of studies and increasing scholarly interest. However, the lack of human-friendly and individual interpretations in explaining the outcomes of machine learning algorithms has significantly hindered the acceptance of these methods by clinicians in their research and clinical practice. To address this issue, our study uses counterfactual explanations to explore the applicability of "what if?" scenarios in medical research. Our aim is to expand our understanding of magnetic resonance imaging (MRI) features used for diagnosing pediatric posterior fossa brain tumors beyond existing boundaries. In our case study, the proposed concept provides a novel way to examine alternative decision-making scenarios that offer personalized and context-specific insights, enabling the validation of predictions and clarification of variations under diverse circumstances. Additionally, we explore the potential use of counterfactuals for data augmentation and evaluate their feasibility as an alternative approach in our medical research case. The results demonstrate the promising potential of using counterfactual explanations to improve AI-driven methods in clinical research.

\keywords{counterfactual explanations \and machine learning \and pediatric brain tumors \and explainable AI \and magnetic resonance imaging}
\end{abstract}
\section{Introduction}

As we incorporate automated decision-making systems into the real world, explainability and accountability questions become increasingly important \cite{mittelstadt2019explaining}. In some fields, such as medicine and healthcare, ignoring or failing to address such a challenge can seriously limit the adoption of computer-based systems that rely on machine learning (ML) and computational intelligence methods for data analysis in real-world applications \cite{vellido2020importance, avanzo2020machine, cinà2022need}. Previous research in eXplainable Artificial Intelligence (XAI) has primarily focused on developing techniques to interpret decisions made by black box ML models. For instance, widely used approaches such as local interpretable model-agnostic explanations (LIME) \cite{ribeiro2016should} and shapley additive explanations (SHAP) \cite{lundberg2017unified} offer attribution-based explanations to interpret ML models. These methods can assist computer scientists and ML experts in understanding the reasoning behind the predictions made by AI models. 

On the other hand, counterfactual explanations \cite{wachter2017counterfactual, miller2019explanation} are a form of model-agnostic interpretation technique that identifies the minimal changes needed in input features to yield a different output, aligned with a specific desired outcome. This approach may be more interesting to end users, including clinicians and patients, rather than focusing solely on how models arrive at their predictions. Because counterfactual explanations can help to better understand the practical consequences of the ML model's predictions and take practical actions at the individual level. Patients' primary concern is not only to learn about their diseases, but also to seek guidance on how to regain their health. For example, counterfactual explanations could potentially be used by clinicians to help inform patients about what small changes they need to make to become healthy again.  
Understanding the doctor's or ML model's decision-making process is less important to patients.

Leveraging counterfactual explanations holds promise in enhancing the data synthesize and for answering causal questions \cite{pawlowski2020deep, sanchez2022healthy}, and interpretability of AI models by offering deeper insights into their decision-making processes \cite{wang2021counterfactual, todo2023counterfactual, nagesh2023explaining}. 
Our proposed approach aims to uncover the underlying reasons for the observed relationships between MRI features, going beyond just generating actionable outcomes for individual patients. Through counterfactual explanations, previously unseen decisions within the decision space can be brought to light. Numerous questions can be explored, such as how to determine the modifications required to transform a patient's diagnosis from one tumor subtype to another. Initially, posing such a question may seem nonsensical and illogical since an individual's actual tumor type cannot be magically altered. However, considering the challenge of distinguishing these two tumor types in clinical settings, asking such a question can effectively demonstrate which features are more informative in differentiating  tumor types.  
Counterfactual explanations enable us to identify the characteristics that distinguish two patient types with the smallest changes in features. Consequently, a deeper understanding of the interactions between MRI features and tumors can be gained; unveiling previously undisclosed outcomes that may be concealed in existing ML studies.

Furthermore, we have identified a potential contribution to clinical practice whereby a new patient with only MRI data available can have their tumor type estimated using a counterfactual approach, prior to receiving histopathological results. Since there is no prior label available for the patient, they are given an "unknown" label and the counterfactual approach is used for each tumor type, allowing estimation of the tumor type with the lowest distance and smallest change in features. While this approach shares similarities with ML, the crucial distinction lies in retaining information about the reasoning behind the estimated tumor type and its corresponding feature changes. This, in turn, can enhance our understanding and the use of AI models in clinical practice.

Last but not least, in situations where the acquisition of data is limited or not possible, various data augmentation methods have been developed to enhance the performance of ML and related applications \cite{wong2016understanding, zhang2017mixup, chawla2002smote}. However, these methods also give rise to additional issues while fulfilling their intended purpose, such as introducing biased shifts in data distribution. To address this issue, we employed counterfactuals generated from different spaces in order to balance the data by maximizing its diversity, and subsequently reported the results for different scenarios.

To summarize, our main contributions include:\begin{itemize}
    \item introducing a new perspective on the application of counterfactual explanations in the pediatric posterior fossa brain tumor literature,
    \item demonstrating how the counterfactual approach, which enables us to provide patient-specific local explanations, can inform us in differentiating tumor types,
    \item providing a systematic comparative analysis of machine explanations, with the aim of evaluating the feasibility of the outcomes.
\end{itemize}

\section{Material \& Methods}
This study integrates a systematic approach combining clinical data acquisition, MRI feature analysis, machine learning, and counterfactual explanations to investigate tumor classification in pediatric posterior fossa brain tumors.

\begin{figure}[htb!]
\begin{center}
\includegraphics[width=\linewidth]{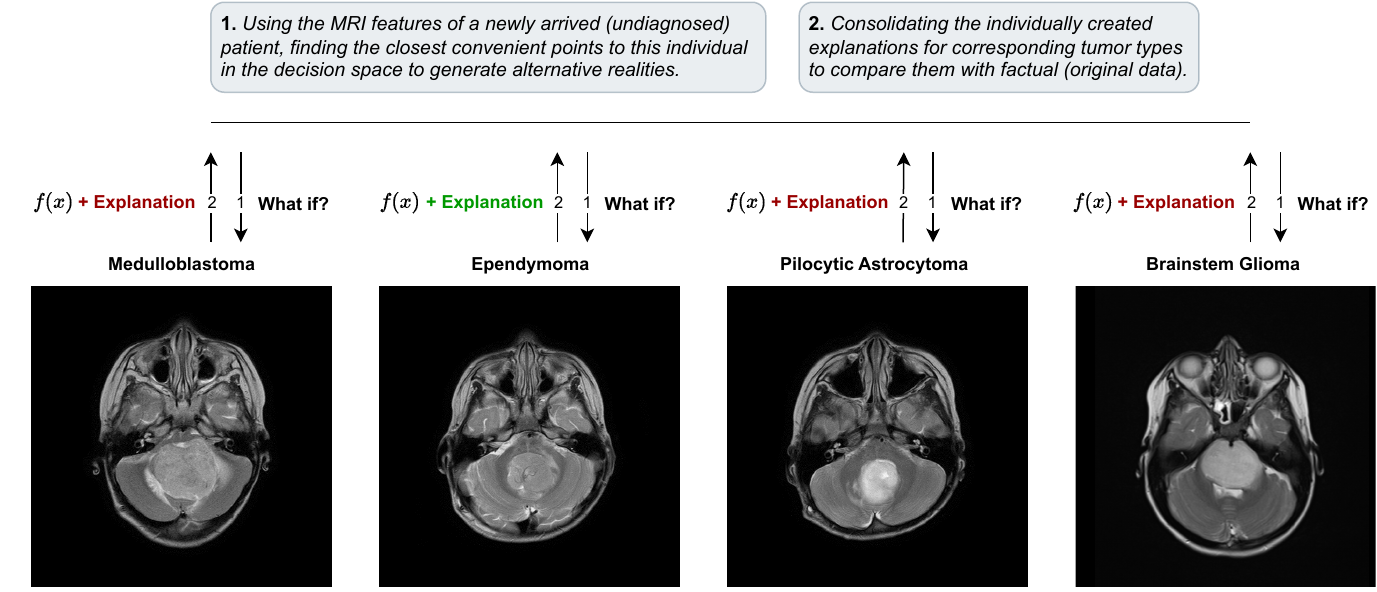}
\caption{The figure illustrates a hypothetical clinical scenario showcasing the practical application of counterfactuals. The data consists of features extracted from provided multi-parametric brain MRIs, rather than the raw images themselves.} \label{Fig:fig2}
\end{center}
\end{figure}

\subsection{Ethics Statement and Patient Characteristics} 
This prospective study (Ref: 632 QÐ-NÐ2 dated 12 May 2019) was carried out in both Radiology and Neurosurgery departments, and was approved by the Institutional Review Board in accordance with the 1964 Helsinki declaration. Written informed consent was obtained from authorized guardians of patients prior to the MRI procedure. Our study comprised a cohort of 112 pediatric patients diagnosed with posterior fossa tumors, including 42 with MB, 25 with PA, 34 with BG, and 11 with EP. All BG patients were confirmed based on full agreement between neuroradiologists and neurosurgeons, whereas the remaining MB, PA, and EP patients underwent either surgery or biopsy for histopathological confirmation.

\subsection{Data Acquisition and Assessment of MRI Features}

For all patients, MRI exams including T1W, T2W, FLAIR, DWI (b values: 0 and 1000) with ADC, and contrast-enhanced T1W (CE-T1) sequences with macrocyclic gadolinium-based contrast enhancement (0.1 ml/kg Gadovist, Bayer, Germany, or 0.2 ml/kg Dotarem, Guerbet, France) were collected in the supine position using a 1.5 Tesla MRI scanner (Multiva, Philips, Best, the Netherlands).

The Medical Imaging Interaction Toolkit (German Cancer Research Center, Division of Medical Image Computing, Heidelberg, Germany) was utilized for measuring the region of interest (ROI) of posterior fossa tumors and normal-appearing parenchyma and and subsequently assessed the following MRI features: signal intensities (SIs) of T2, T1, FLAIR, T1CE, DWI, and ADC. Ratios between the posterior fossa tumor and parenchyma were calculated by dividing the SI of the tumor and the SI of the normal-appearing parenchyma based on T2, T1, FLAIR, T1CE, DWI, and ADC. Additionally, ADC values were quantified for both the posterior fossa tumor and parenchyma on the ADC map using the MR Diffusion tool available in Philips Intellispace Portal, version 11 (Philips, Best, The Netherlands). It is worth noting that, prior to analysis, bias field correction was applied to every image to correct for nonuniform grayscale intensities in the MRI caused by field inhomogeneities.

\subsection{Standardization} 

Prior to conducting ML trainings, the dataset was subjected to a standardization process, using Python programming (version 3.9.13) with the Scikit-Learn library (version 1.0.2) module. This technique involved transforming the data to have a mean of zero and a standard deviation of one. To standardize all numerical attributes, the Scikit-Learn StandardScaler function was employed, which subtracted the mean and scaled the values to unit variance, ensuring the data was in a standardized format. To determine the standard score of a sample $x_i$ , the following formula is used: 
\begin{equation}
z = \frac{x_i - \mu}{\sigma},
\end{equation}
where, $\mu$ represents the mean of the training samples, and $\sigma$ represents their standard deviation.

\subsection{Distance Calculation}

Utilizing counterfactuals as classifiers, the notable discrepancy in MRI feature values, as shown in the example in Table \ref{table2}, complicates distance calculations.
The underlying reason for the distance calculation issue is that some MRI features in fact include ratios that depend on other variables. We addressed this by omitting unchanged values (`-'), rescaling the rest to a consistent scale, and then reintroducing them. The distances were then computed using the Euclidean metric on the counterfactual values of the current factual. Minimizing this distance aids in determining the tumor type by corresponding to the least dissimilarity (Table \ref{table4}). The Euclidean distance is given by:

\begin{equation}
    \mathrm{Distance} = \sqrt{\sum_{i=1}^{n} (x_i - y_i)^2}
\end{equation}

Here, $x_i$ and $y_i$ are values from the current and baseline rows, respectively. The formula sums the squared differences for each feature, then takes the square root. In this equation, $n$ signifies the feature count in the dataset.

\subsection{Statistical Analysis}\label{statisticalAn}

Using the $t$-test from the \texttt{scipy} library (version 1.10.1), we considered a two-tailed $p$-value of $<0.05$ as statistically significant. Our analysis comprised:
\begin{enumerate}
    \item Assessing the statistical significance of changes in counterfactuals when transitioning the tumor type from $\mathcal{X}$ to $\mathcal{Y}$ (dependent $t$-test).
    \item Comparing the similarity between counterfactuals transitioning from $\mathcal{X}$ to $\mathcal{Y}$ and the original patients with tumor type $\mathcal{Y}$ (Welch's $t$-test).
\end{enumerate}

For each transition from tumor type $\mathcal{X}$ to $\mathcal{Y}$, we generated five counterfactuals. In our $t$-test evaluations, these counterfactuals were examined in distinct manners. Due to generating a dataset larger than our original sample size, we could not maintain equal dimensions for the dependent analysis. To address this:
\begin{itemize}
    \item For each counterfactual, we used the data of the corresponding factual patient as a baseline for testing.
    \item We subsequently tested the significance of the top five most changed feature variables.
\end{itemize}

For the independent analysis, all counterfactuals were evaluated using the real data of all patients, focusing on the three most altered features. Each feature was evaluated separately.

In summary, the primary distinction between our tests was their focus: one on patients, and the other on features.

\subsection{Distribution Plotting}
To generate individual kernel density estimation (KDE) plots for each feature, we utilized the kdeplot function from the Seaborn package (version 0.11.2). By specifying a hue parameter (e.g., Tumor Type), we were able to incorporate a meaningful association using this method. Consequently, we transformed the default marginal plot into a layered KDE plot. This approach tackles the challenge of reconstructing the density function $f$ using an independent and identically distributed (iid) sample $x_1, x_2,..., x_n$ from the respective probability distribution.

\subsection{Machine Learning}

To decrease overfitting and convergence issue of counterfactuals, especially for EP, we took less patients to implement the task: 25 patients from MB, PA and BG and 11 patients from EP. For testing, to ensure the reliability of our ML models, particularly with a small dataset, we conducted five runs using stratified random sampling based on tumor type with 55\% train and 45\% test patients. 

Using nine ML models, including support vector machine (SVM), adaboost (ADA), logistic regression (LR), random forest classifier (RF), decision tree classifier (DT), gradient boosting classifier (GB), catboost classifier (CB), extreme gradient boosting classifier (XGB) and voting classifier (VOTING), we evaluated the models on the raw data with the outcomes prior to our counterfactual interpretations. CB and XGB were obtained from CatBoost version 1.1.1 and XGBoost version 1.5.1 libraries, respectively, while the other models were obtained from the Scikit-Learn library.

We assessed the performance of the models using precision, recall, and F1 score, which were calculated based on the counts of true positives (TP), true negatives (TN), false positives (FP), and false negatives (FN). In order to ensure an accurate interpretation of the ML results, we opted not to balance the labels. Instead, we employed macro precision, macro recall, and macro F1 score metrics, which take into account the contributions of all labels equally. This approach enabled us to observe the genuine impact of the varying label frequencies, EP in this case.

\section{Counterfactual Explanations}

Given the challenges associated with local approximations, it is worthwhile to explore prior research in the "explanation sciences" to identify potential alternative strategies for generating reliable and practical post-hoc interpretations that benefit the stakeholders affected by algorithmic decisions \cite{mittelstadt2019explaining, ruben2015explaining}. To create explanations that are understandable and useful for both experts and non-experts, it is logical to investigate theoretical and empirical studies that shed light on how humans provide and receive explanations \cite{miller2019explanation}. Over the past few decades, the fields of philosophy of science and epistemology have shown increasing interest in theories related to counterfactual causality and contrastive explanations \cite{ruben2015explaining, Lewis1973-LEWC-2, kment2006counterfactuals, woodward1997explanation, woodward2017scientific}.

In philosophy, counterfactuals serve not only to assess the relationship between a mental state and reality, but also to determine whether a mental state can be considered as knowledge. The problem of identifying knowledge with justified true belief is complicated by various counterexamples, such as Gettier cases (1963) \cite{Gettier1963-GETIJT-4}. However, some scholars proposed additional conditions to address these counterexamples. This literature highlighted two significant counterfactual conditions:
\par \textbf{Sensitivity:} If $\rho$ were false, $\mathcal{S}$ would not believe that $\rho$.
\par \textbf{Safety:} If $\mathcal{S}$ were to believe that $\rho$, $\rho$ would not be false.

Both of these conditions express the notion that $\mathcal{S}$'s beliefs must be formed in a manner that is sensitive to the truthfulness of $\rho$. The counterfactual semantics has influenced from this idea in various ways, including the establishment of their non-equivalence, clarification, and resolution of potential counterexamples \cite{starr2019counterfactuals}.

This concept has sparked a fresh wave of counterfactual analyses that employ new methodologies. Hitchcock \cite{hitchcock2001intransitivity, hitchcock2007prevention} and Woodward \cite{woodward2005making}, for instance, constructed counterfactual analyses of causation using Bayesian networks (also known as "causal models") and structural equations. The basic idea of the analysis can be summarized as follows: "$\mathcal{X}$ can be considered a cause of $\mathcal{Y}$ only if there exists a path from $\mathcal{X}$ to $\mathcal{Y}$, and changing the value of $\mathcal{X}$ alone results in a change in the value of $\mathcal{Y}$".

Ginsberg (1986) \cite{ginsberg1986counterfactuals} initiated his discussion by outlining the potential significance of counterfactuals in artificial intelligence and summarizing the philosophical insights that have been drawn regarding them. Following this, Ginsberg provided a structured explanation of counterfactual implication and analyzed the challenges involved in executing it. Over time, numerous developments in the fields of artificial intelligence and cognitive science, including the Bayesian epistemology approach, have gone beyond what was previously envisioned by Ginsberg regarding the potential application of artificial intelligence and counterfactuals \cite{wachter2017counterfactual, miller2019explanation, spirtes1993discovery, spirtes2000causation, griffiths2010probabilistic, chou2022counterfactuals}. Furthermore, Verma et al. \cite{verma2020counterfactual} conducted a comprehensive review of the counterfactual literature, analyzing its utilization in over 350 research papers.

In recent times, there has been a growing interest in the concept of counterfactual explanations, which aim to provide alternative perturbations capable of changing the predictions made by a model. In simple terms, when given an input feature $x$ and the corresponding output produced by an ML model $f$, a counterfactual explanation involves modifying the input to generate a different output $y$ using the same algorithm. To further explain this concept, Wachter et al. \cite{wachter2017counterfactual} introduce the following formulation in their proposal:
\begin{equation}
    c = \arg \min\limits_{c} \ell(f(c), y) + |x - c|
\end{equation}
The initial component $\ell$ of the formulation encourages the counterfactual $c$ to deviate from the original prediction, aiming for a different outcome. Meanwhile, the second component ensures that the counterfactual remains in proximity to the original instance, thereby emphasizing the importance of maintaining similarity between the two.

While challenges like the inability to find optimal counterfactual explanations underscore the need for DiCE updates, there are alternative solutions. Dutta et al. \cite{dutta2022robust} and Maragno et al. \cite{maragno2023finding} propose alternative counterfactual algorithms to potentially overcome these challenges. Furthermore, Guidotti's review \cite{guidotti2022counterfactual} presents an extensive list of counterfactual algorithms.

Although the subject of explainable artificial intelligence has been investigated by many researchers in the field of medicine, there are not many studies exploiting counterfactual explanations \cite{borys2023explainable,de2023explainable}. Among explainable AI methods, it appears that SHAP, Grad-CAM, and Lime have been extensively studied \cite{tjoa2020survey,band2023application}. Sarp et al. focused on a LIME-based heat map application for interpretation of COVID-19 cases from chest X-ray images \cite{sarp2023xai}. In another study \cite{lin2022sspnet}, authors applied gradient-weighted class activation mapping (Grad-CAM) to localize decision-making regions. Knapič et al. investigated the use of SHapley Additive exPlanations (SHAP) in medical images and compared its performance with LIME and the Contextual Importance and Utility (CIU) method \cite{knapivc2021explainable}. In \cite{zeineldin2022explainability}, the authors explored the use of different XAI methods including Vanilla gradient, guided backpropagation, integrated gradients, guided integrated gradients, SmoothGrad, Grad-CAM, and guided Grad-CAM to explain brain tumor segmentation. Their work focused on methods for creating visualization maps to better interpret deep learning models.  
Differently in this study, we propose a counterfactual explanation-based approach that explores alternative decision-making scenarios to provide personalized and context-specific insights based on MRI data.

\subsection{Generating Counterfactual Explanations}\label{generateCF}

The clear interpretability of counterfactuals helps an individual make decisions about his or her future  \cite{rockoff2011can, waters2014grade, dai2015prediction, andini2017targeting, athey2017beyond, yang2020generating, wang2021counterfactual, xu2022counterfactual, yang2023counterfactual}. In situations where a \textit{negative} response is received, understanding how to improve results without resorting to major and unrealistic data alterations becomes important.

\begin{figure}[htb!]
\begin{center}
\includegraphics[width=\linewidth]{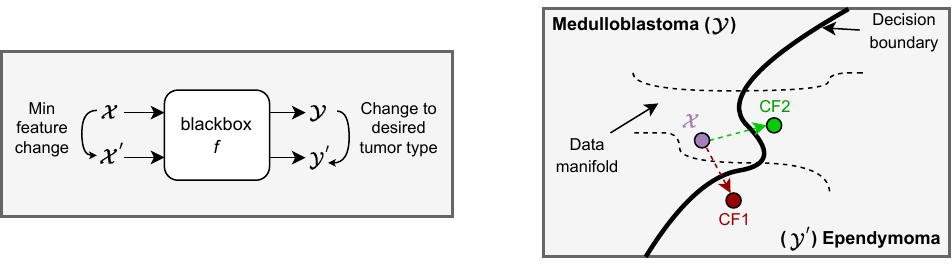}
\caption{Generating counterfactual explanations on tumor types. The left illustrates feature manipulation using an ML model with a counterfactual approach. The right delves deeper into the process and counterfactual explanation concept, exemplified by two tumor types.}
\label{Fig:fig1}
\end{center}
\end{figure}

We argue that the use of counterfactual explanations effectively leverages the factual insights derived from MRI features. These features act as distinct markers, facilitating the differentiation between various tumors. Such a methodology shines particularly when traditional diagnostics find it challenging to distinguish between tumor types. 

The \textit{data manifold} concept, illustrated in Fig. \ref{Fig:fig1}, emphasizes the importance of proximity in counterfactual explanations. For counterfactuals to be credible, their features should resemble those of prior classifier observations and be realistic. Counterfactuals with features that diverge from training data or disrupt feature associations are impractical and outside established norms \cite{brown2021uncertainty}. To ensure that counterfactuals are realistic and align with the training data, we employ constraint-based approaches in our algorithms. For instance, changing parameters such as "age" and "gender" would be highly unreasonable. Therefore, in most scenarios, these parameters are specified in the model to prevent counterfactuals from deviating from reality. Similarly, in our case, while parenchyma features were included during training, they remain invariant during counterfactual generation to preserve tissue characteristic references.

The DiCE library \cite{mothilal2020explaining} offers a framework for counterfactual generation, viewing it as an optimization task akin to adversarial example discovery. Crucially, the modifications must be diverse, feasible, and implementable. The optimization formula used is:
\begin{equation}
    C(x) = \arg\min\limits_{c_1,...,c_k} \frac{1}{k}\sum_{i=1}^k \ell(f(c_i), y) + \frac{\lambda_1}{k} \sum_{i=1}^k \mathrm{dist}(c_i, x) - \lambda_2 \, \mathrm{dpp\_diversity}(c_1,...,c_k),
\end{equation}
with hinge loss as:
\begin{equation}
    \ell = \max(0, 1 - z * \mathrm{logit}(f(c))),
\end{equation}
where $z$ takes values based on $y$ and logit($f(c)$) represents unscaled ML output. Diversity is expressed as:
\begin{equation}
    \mathrm{dpp\_diversity} = \mathrm{det(K)},
\end{equation}
where $\mathrm{K}_{i,j} = \frac{1}{1+\mathrm{dist}(c_i,c_j)}$, and the distance between counterfactuals is measured.

\section{Results}

\subsection{What if the counterfactual explanations graciously provide us with additional insights into classification?}

\begin{figure}[htb!]
\begin{center}
\includegraphics[width=\linewidth]{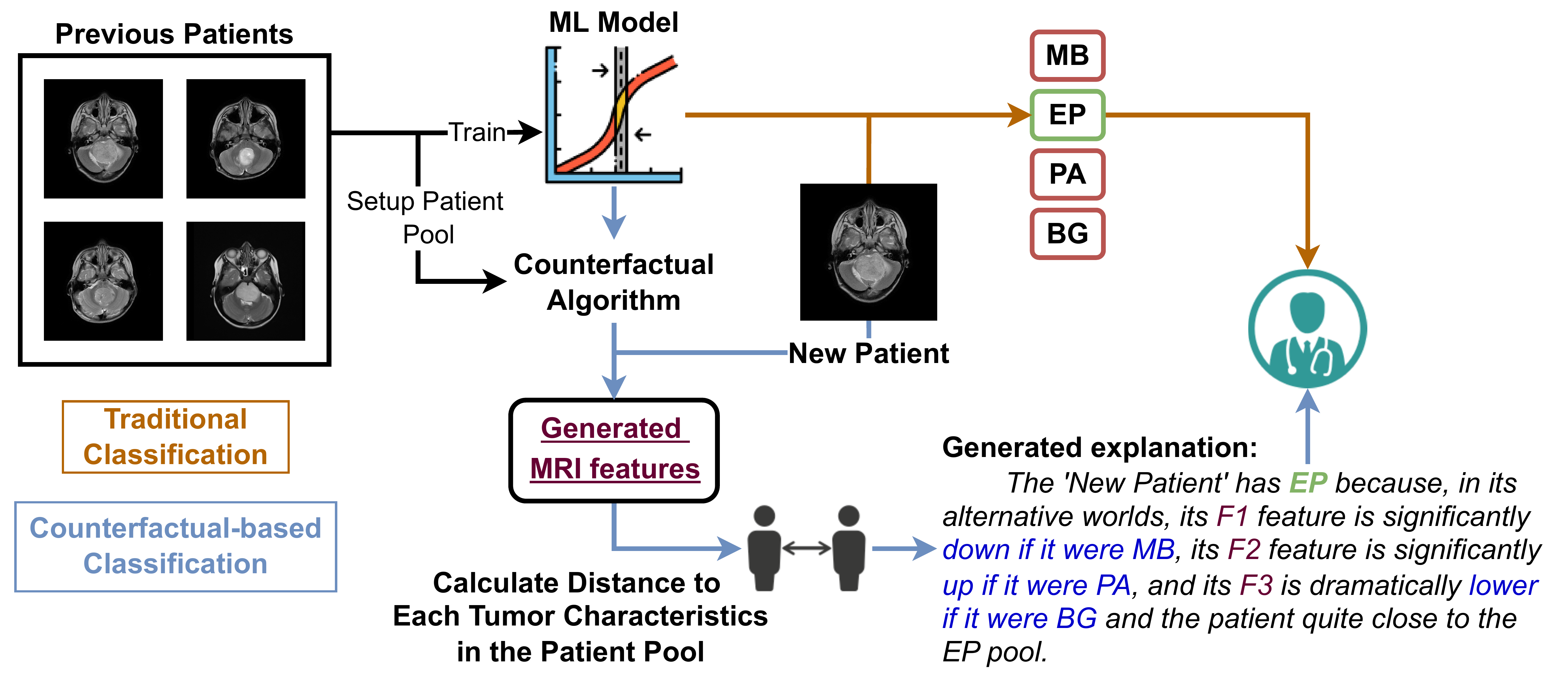}
\caption{An overview of our approach demonstrating the generation of counterfactual explanations.}
\label{Fig:fignew1}
\end{center}
\end{figure}

Using DiCE's multi-class training capability, we have established a framework that simultaneously trains for four distinct tumor types. This framework employs counterfactual explanations, serving as classifiers, to determine tumor types based on numeric MRI data. Figure \ref{Fig:fignew1} presents an overview of our approach demonstrating the generation of counterfactual explanations. By utilizing all four tumor types, we essentially construct a decision space of reality with our existing patients. As the new patient is guided through this space, attempting to transform into each disease sequentially, the degree of self-modification required for each specific tumor condition will vary. As the required changes decrease, it can be inferred that the patient is closer to that particular tumor type since they necessitate fewer modifications. Similarly, understanding the level of dissimilarity and the contributing features to this dissimilarity has been explored as a critical approach in determining the tumor type. In our previous study \cite{tanyel4421493deciphering} 
the LR model excelled in binary classification and generating counterfactuals, thereby justifying its selection for our current research.

Figure \ref{Fig:fig2} and Table \ref{table2} present a scenario involving a patient diagnosed via MRI with an indeterminate tumor type. Using the MRI data, we generate "what-if" scenarios for each tumor type. These scenarios, based on feature distances, help determine the tumor type the patient's data aligns with.

\renewcommand{\arraystretch}{1.25}
\begin{table}[htb!]
\centering
\resizebox{\textwidth}{!}{%
\begin{tabular}{ccccccccccccc}
{\color[HTML]{1155CC} \textbf{Factual ($x$)}} &
   &
   &
   &
   &
   &
   &
   &
   &
   &
   &
   &
   \\ \cline{1-1}
\textbf{Tumor Type} &
  \textbf{T2\_T} &
  \textbf{T2\_R} &
  \textbf{FLAIR\_T} &
  \textbf{FLAIR\_R} &
  \textbf{DWI\_T} &
  \textbf{DWI\_R} &
  \textbf{ADC\_T} &
  \textbf{ADC\_R} &
  \textbf{T1\_T} &
  \textbf{T1\_R} &
  \textbf{T1CE\_T} &
  \textbf{T1CE\_R} \\ \hline
\textit{unknown (EP)} &
  1286 &
  1.529 &
  1311 &
  1.341 &
  1175 &
  1.088 &
  1.009 &
  1.771 &
  473 &
  0.84 &
  892 &
  1.595
\end{tabular}}
\vspace{5pt} \label{original1}
\centering
\resizebox{\textwidth}{!}{%

\begin{tabular}{ccccccccccccc}
\multicolumn{3}{l}{{\color[HTML]{6200C9} \textbf{Counterfactual ($x_{cf}$)}}} &     &   &        &   &   &   &   &   &        &       \\ \cline{1-3}
\textbf{Tumor Type} &
  \textbf{T2\_T} &
  \textbf{T2\_R} &
  \textbf{FLAIR\_T} &
  \textbf{FLAIR\_R} &
  \textbf{DWI\_T} &
  \textbf{DWI\_R} &
  \textbf{ADC\_T} &
  \textbf{ADC\_R} &
  \textbf{T1\_T} &
  \textbf{T1\_R} &
  \textbf{T1CE\_T} &
  \textbf{T1CE\_R} \\ \hline
MB                     & -                          & -                    & 648 & - & -      & - & - & - & - & - & 1309.5 & -     \\
EP                     & 1423.2                     & -                    & -   & - & -      & - & - & - & - & - & -      & -     \\
PA                     & 2290.2                     & -                    & -   & - & -      & - & 2 & - & - & - & 1492.5 & -     \\
BG                     & -                          & -                    & -   & - & 544.23 & - & 2 & - & - & - & -      & 0.781
\end{tabular}}
\vspace{5pt}
\caption{This table presents the results of our proposed method utilizing counterfactuals (Fig. \ref{Fig:fig2}).} \label{table2}
\end{table}

Table \ref{table2} details an unknown patient classified as EP. For the MB counterfactual, changes in FLAIR\_Tumor and T1CE\_Tumor result in distances of -663 and 417.5, respectively. For EP, T2\_Tumor changes, resulting in a distance of 137.2. In the PA group, changes are observed in T2\_Tumor (from 1286 to 2290.2), ADC\_Tumor (from 1.009 to 2), and T1CE\_Tumor (from 892 to 1492.5). For the BG group, DWI\_Tumor changes from 1175 to 544.23, ADC\_Tumor from 1.009 to 2, and T1CE\_Ratio from 1.595 to 0.781.

In Tables \ref{table2} and \ref{table4}, T represents Tumor and R represents Ratio (Tumor/Parenchyma). The symbol (-) indicates no modification. In Table \ref{table2}, the patient, with the lowest feature distance to EP, is predicted as EP.

Table \ref{table4} displays four new samples after standardization of features and distance calculation. Distances are provided for each patient's tumor type and their respective counterfactuals. The data represented by (-) matches original values. The distance metric adjusts the distance magnitude between them.

\renewcommand{\arraystretch}{1.3}
\begin{table}[htb!]
\centering
\resizebox{\textwidth}{!}{%
\begin{tabular}{ccccccccccccccc}
\textbf{} &
  \textbf{Tumor Type} &
  \textbf{T2\_T} &
  \textbf{T2\_R} &
  \textbf{FLAIR\_T} &
  \textbf{FLAIR\_R} &
  \textbf{DWI\_T} &
  \textbf{DWI\_R} &
  \textbf{ADC\_T} &
  \textbf{ADC\_R} &
  \textbf{T1\_T} &
  \textbf{T1\_R} &
  \textbf{T1CE\_T} &
  \textbf{T1CE\_R} &
  \multicolumn{1}{l}{\textbf{Distance}} \\ \hline
\textit{Factual ($x$)}    & \textit{unknown (MB)} & 0      & -0.5   & -0.5   & 0.5    & 0   & 0      & -0.5   & -1.191 & 0    & 0   & 0      & 0    & -     \\
Counterfactual ($x_{cf}$) & MB                    & -      & -      & -      & -2     & -   & -      & -      & -      & -    & -   & -      & -    & \textbf{2.5}   \\
Counterfactual ($x_{cf}$) & EP                    & -      & -      & 2      & -      & -   & -      & -      & 0.370  & -    & -   & -      & -    & 2.948 \\
Counterfactual ($x_{cf}$) & PA                    & -      & 2      & -      & -      & -   & -      & -      & 0.993  & -    & -   & -      & -    & 3.319 \\
Counterfactual ($x_{cf}$) & BG                    & -      & -      & -      & -      & -   & -      & 2      & 1.020  & -    & -   & -      & -    & 3.337 \\ \hline
\textit{Factual ($x$)}    & \textit{unknown (EP)} & -0.583 & 0      & 0.5    & 0      & 0.5 & 0      & -0.816 & 0      & 0    & 0   & -0.795 & 0.5  & -     \\
Counterfactual ($x_{cf}$) & MB                    & -      & -      & -2     & -      & -   & -      & -      & -      & -    & -   & 0.836  & -    & 2.985 \\
Counterfactual ($x_{cf}$) & EP                    & -0.233 & -      & -      & -      & -   & -      & -      & -      & -    & -   & -      & -    & \textbf{0.350} \\
Counterfactual ($x_{cf}$) & PA                    & 1.982  & -      & -      & -      & -   & -      & 1.225  & -      & -    & -   & 1.550  & -    & 4.031 \\
Counterfactual ($x_{cf}$) & BG                    & -      & -      & -      & -      & -2  & -      & 1.225  & -      & -    & -   & -      & -2   & 4.082 \\ \hline
\textit{Factual ($x$)}    & \textit{unknown (PA)} & 0.5    & 1.223  & 0      & 0.5    & 0.5 & 0.124  & 0.816  & 0      & 0    & 0.5 & 0      & 0.5  & -     \\
Counterfactual ($x_{cf}$) & MB                    & -      & -0.871 & -      & -2     & -   & -      & -1.225 & -      & -    & -2  & -      & -    & 4.588 \\
Counterfactual ($x_{cf}$) & EP                    & -2     & -0.869 & -      & -      & -2  & -1.749 & -1.225 & -      & -    & -   & -      & -    & 4.955 \\
Counterfactual ($x_{cf}$) & PA                    & -      & -      & -      & -      & -   & 1.377  & -      & -      & -    & -   & -      & -    & \textbf{1.252} \\
Counterfactual ($x_{cf}$) & BG                    & -      & -0.705 & -      & -      & -   & -      & -      & -      & -    & -   & -      & -2   & 3.157 \\ \hline
\textit{Factual ($x$)}    & \textit{unknown (BG)} & 0.811  & 0.5    & -0.373 & 0.811  & 0   & 0      & 0.831  & 0.5    & -0.5 & 0.5 & -0.602 & -0.5 & -     \\
Counterfactual ($x_{cf}$) & MB                    & -1.033 & -      & -0.847 & -1.393 & -   & -      & -1.079 & -2     & 2    & -2  & -0.166 & 2    & 6.109 \\
Counterfactual ($x_{cf}$) & EP                    & -1.400 & -2     & 1.966  & -      & -   & -      & -1.360 & -      & -    & -   & -      & -    & 4.627 \\
Counterfactual ($x_{cf}$) & PA                    & -      & -      & -      & -      & -   & -      & 0.778  & -      & -    & -   & 1.971  & -    & 2.574 \\
Counterfactual ($x_{cf}$) & BG                    & -      & -      & -      & -1.041 & -   & -      & -      & -      & -    & -   & -      & -    & \textbf{1.853}
\end{tabular}}
\vspace{5pt}
\caption{Distance results for counterfactuals generated on feature-wise scaled data for four distinct newly arriving patients with varying tumor types.}\label{table4}
\end{table}

\subsection{Revealing Key MRI Features through Counterfactual Explanations}

As discussed in Section \ref{generateCF}, counterfactual explanations can provide insights into feature importance. These explanations allow us to understand the reasoning behind ML model decisions and offer valuable options for restriction. In clinical settings, visible changes in features through counterfactual explanations can be more relevant and meaningful for real-world evaluations and applications.

Considering that we generated five counterfactuals for each patient, we obtained 125 explanations for MB, PA, and BG, and 55 explanations for EP. Table \ref{table5} illustrates our reporting method for counterfactual analysis results for a case scenario (MB to EP). The patient count, the total number of generated counterfactual explanations for them, and the statistical information regarding the frequency of changes observed on which features in these counterfactuals to identify the top 3 influential features are shown. For instance, "FLAIR\_Tumor 71 changes" signifies that out of 125 counterfactuals, 71 of them involved a modification from MB to EP. Therefore, FLAIR\_Tumor creates such a distinction between these two tumors that the model considers altering this feature significantly influential in shifting the decision from one side to the other in the decision space. The greater the repetition of this occurrence, indicated by the magnitude of "changes," the more pronounced the outcome suggesting that even in random selections, optimization is achieved for that particular feature, significantly impacting the decision.
\renewcommand{\arraystretch}{1.3}
\begin{table}[htb!]
\centering
\resizebox{0.7\textwidth}{!}{%
\begin{tabular}{lcllc}
\multicolumn{5}{l}{Number of patients: 25}                   \\
\multicolumn{5}{l}{Number of generated counterfactuals: 125} \\ \hline
FLAIR\_Tumor & 71 changes &  & T1\_Ratio         & 6 changes \\
ADC\_Tumor   & 33 changes &  & T1CE\_Tumor       & 6 changes \\
ADC\_Ratio   & 29 changes &  & T2\_Tumor         & 3 changes \\
DWI\_Ratio   & 18 changes &  & T2\_Parenchyma    & 0 changes \\
FLAIR\_Ratio & 17 changes &  & FLAIR\_Parenchyma & 0 changes \\
DWI\_Tumor   & 12 changes &  & DWI\_Parenchyma   & 0 changes \\
T1\_Tumor    & 10 changes &  & ADC\_Parenchyma   & 0 changes \\
T1CE\_Ratio  & 7 changes  &  & T1\_Parenchyma    & 0 changes \\
T2\_Ratio    & 6 changes  &  & T1CE\_Parenchyma  & 0 changes
\end{tabular}}\vspace{0.2cm}
\caption{This example analysis presents the variations in characteristics observed during the generation of counterfactual instances for the transition from MB to EP.}\label{table5}
\end{table}
Table \ref{best3} presents the findings from each tumor pair to identify feature differences between different tumor types. The observed changes in features align with expected outcomes from clinical studies. MB and EP tumors are distinguished by FLAIR and ADC features. MB and PA typically exhibit differences in T2 and ADC. MB and BG, on the other hand, show variations primarily in ADC, T2, and T1CE. In the case of EP and PA, T2 exhibits the most significant changes, while variations in ADC and T1CE are also observed. The most distinguishing features between EP and BG are T1CE\_Ratio and ADC\_Tumor. As for PA and BG, the T2\_Ratio feature has been identified as a crucial factor in creating differentiation. Additionally, significant variations in T1CE features are frequently observed, further contributing to the dissimilarity between these tumor types.

\begin{figure}[htbp]
\begin{center}
\includegraphics[width=\linewidth]{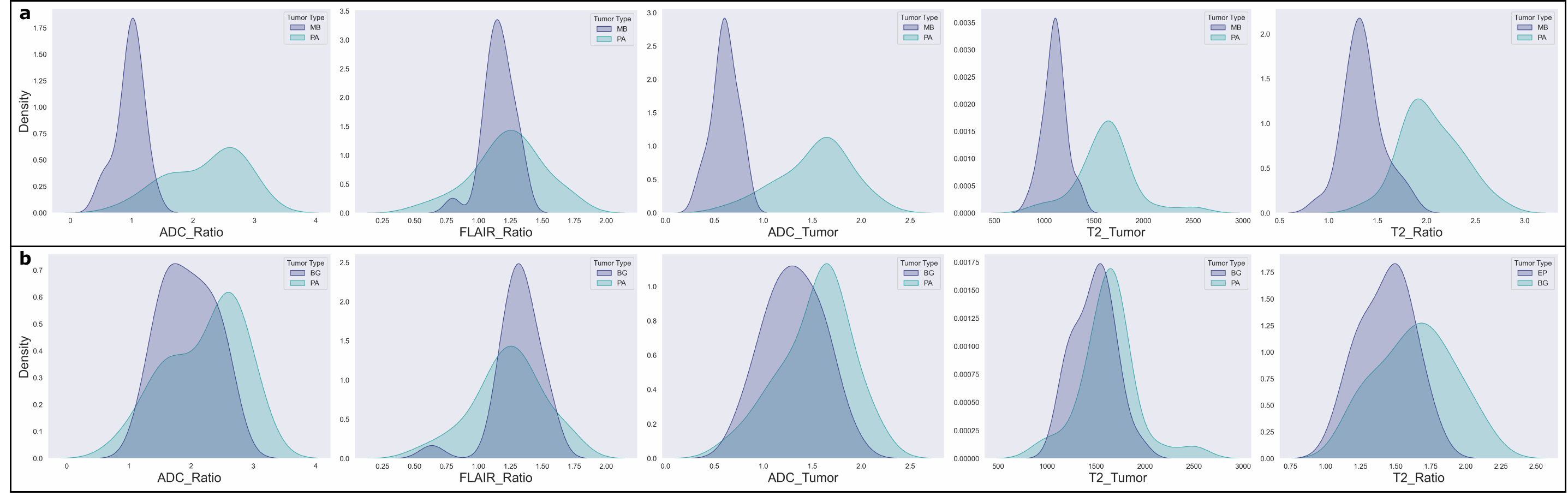}
\caption{The original data distributions for MB-PA and BG-PA, highlighting key features for MB-PA and their behavior when applied to BG-PA. The figure demonstrates how the top 5 features vary significantly between MB and PA, while BG and PA distributions remain almost identical, lacking discernible features.
} 
\label{Fig:Fig3}
\end{center}
\end{figure}

The results illustrated in Figure
\ref{Fig:Fig3} provide a clear view of the original data distributions for MB-PA and BG-PA, shedding light on critical MRI features through counterfactual explanations. This figure, along with Table \ref{best3}, reveals that transitions between MB and PA generally exhibit similar overall distributions. However, there are significant differences in the top 5 features, which aligns with our previous study \cite{tanyel4421493deciphering}. In contrast, the distributions for BG and PA are nearly identical, with no distinct features. This lack of variation is also evident in the top 5 features, demonstrating the effectiveness of our algorithm in generating counterfactuals by selecting features that show the most variation, thus achieving maximum impact with minimal changes. Notably, T1CE features and T2\_Ratio stand out as the most distinctive between PA and BG, as previously highlighted in Figure 5 of our earlier research \cite{tanyel4421493deciphering}.

\renewcommand{\arraystretch}{1.3}
\begin{table}[htb!]
\centering
\resizebox{\textwidth}{!}{%
\begin{tabular}{lcllcllcllcllcllc}
\multicolumn{2}{c}{\textbf{MB to EP}} &
  \multicolumn{1}{c}{\textbf{}} &
  \multicolumn{2}{c}{\textbf{MB to PA}} &
  \multicolumn{1}{c}{\textbf{}} &
  \multicolumn{2}{c}{\textbf{MB to BG}} &
   &
  \multicolumn{2}{c}{\textbf{EP to MB}} &
  \multicolumn{1}{c}{\textbf{}} &
  \multicolumn{2}{c}{\textbf{EP to PA}} &
  \multicolumn{1}{c}{\textbf{}} &
  \multicolumn{2}{c}{\textbf{EP to BG}} \\ \cline{1-2} \cline{4-5} \cline{7-8} \cline{10-11} \cline{13-14} \cline{16-17} 
\multicolumn{1}{c}{\textbf{Feature}} &
  \textbf{Change} &
  \multicolumn{1}{c}{\textbf{}} &
  \multicolumn{1}{c}{\textbf{Feature}} &
  \textbf{Change} &
  \multicolumn{1}{c}{\textbf{}} &
  \multicolumn{1}{c}{\textbf{Feature}} &
  \textbf{Change} &
   &
  \multicolumn{1}{c}{\textbf{Feature}} &
  \textbf{Change} &
  \multicolumn{1}{c}{\textbf{}} &
  \multicolumn{1}{c}{\textbf{Feature}} &
  \textbf{Change} &
  \multicolumn{1}{c}{\textbf{}} &
  \multicolumn{1}{c}{\textbf{Feature}} &
  \textbf{Change} \\ \cline{1-2} \cline{4-5} \cline{7-8} \cline{10-11} \cline{13-14} \cline{16-17} 
FLAIR\_Tumor &
  71 &
  \multicolumn{1}{c}{} &
  T2\_Ratio &
  87 &
  \multicolumn{1}{c}{} &
  T2\_Tumor &
  64 &
   &
  T1CE\_Tumor &
  18 &
   &
  T2\_Tumor &
  34 &
   &
  ADC\_Tumor &
  22 \\
ADC\_Tumor &
  33 &
   &
  T2\_Tumor &
  55 &
   &
  ADC\_Tumor &
  52 &
   &
  FLAIR\_Ratio &
  16 &
   &
  T2\_Ratio &
  26 &
   &
  DWI\_Ratio &
  16 \\
ADC\_Ratio &
  29 &
   &
  ADC\_Tumor &
  43 &
   &
  T1CE\_Ratio &
  43 &
   &
  FLAIR\_Tumor &
  13 &
   &
  ADC\_Tumor &
  19 &
   &
  T1CE\_Ratio &
  15 \\
 &
  \multicolumn{1}{l}{} &
   &
   &
  \multicolumn{1}{l}{} &
   &
   &
  \multicolumn{1}{l}{} &
   &
   &
  \multicolumn{1}{l}{} &
   &
   &
  \multicolumn{1}{l}{} &
   &
   &
  \multicolumn{1}{l}{} \\
\multicolumn{2}{c}{\textbf{PA to MB}} &
  \multicolumn{1}{c}{\textbf{}} &
  \multicolumn{2}{c}{\textbf{PA to EP}} &
  \multicolumn{1}{c}{\textbf{}} &
  \multicolumn{2}{c}{\textbf{PA to BG}} &
  \multicolumn{1}{c}{\textbf{}} &
  \multicolumn{2}{c}{\textbf{BG to MB}} &
  \multicolumn{1}{c}{\textbf{}} &
  \multicolumn{2}{c}{\textbf{BG to EP}} &
  \multicolumn{1}{c}{\textbf{}} &
  \multicolumn{2}{c}{\textbf{BG to PA}} \\ \cline{1-2} \cline{4-5} \cline{7-8} \cline{10-11} \cline{13-14} \cline{16-17} 
\multicolumn{1}{c}{\textbf{Feature}} &
  \textbf{Change} &
  \multicolumn{1}{c}{\textbf{}} &
  \multicolumn{1}{c}{\textbf{Feature}} &
  \textbf{Change} &
  \multicolumn{1}{c}{\textbf{}} &
  \multicolumn{1}{c}{\textbf{Feature}} &
  \textbf{Change} &
   &
  \multicolumn{1}{c}{\textbf{Feature}} &
  \textbf{Change} &
  \multicolumn{1}{c}{\textbf{}} &
  \multicolumn{1}{c}{\textbf{Feature}} &
  \textbf{Change} &
  \multicolumn{1}{c}{\textbf{}} &
  \multicolumn{1}{c}{\textbf{Feature}} &
  \textbf{Change} \\ \cline{1-2} \cline{4-5} \cline{7-8} \cline{10-11} \cline{13-14} \cline{16-17} 
ADC\_Ratio &
  91 &
   &
  T2\_Ratio &
  95 &
   &
  T2\_Ratio &
  95 &
   &
  FLAIR\_Ratio &
  85 &
   &
  T1CE\_Ratio &
  90 &
   &
  T1CE\_Ratio &
  53 \\
FLAIR\_Ratio &
  88 &
   &
  T2\_Tumor &
  81 &
   &
  T1CE\_Tumor &
  66 &
   &
  ADC\_Tumor &
  71 &
   &
  T1\_Tumor &
  50 &
   &
  T1CE\_Tumor &
  48 \\
ADC\_Tumor &
  76 &
   &
  T1CE\_Tumor &
  45 &
   &
  T1CE\_Ratio &
  54 &
   &
  ADC\_Ratio &
  71 &
   &
  DWI\_Ratio &
  48 &
   &
  T2\_Ratio &
  32
\end{tabular}}\vspace{0.2cm}
\caption{The three most important features for each changing reality case.}\label{best3}
\end{table}

\subsection{Statistical Analysis of Generated Counterfactuals}\label{section4-3}

\renewcommand{\arraystretch}{1.3}
\begin{table}[b!]
\centering
\label{MB}
\resizebox{0.482\textwidth}{!}{%
\subfloat[Difference between original MB and generated MBs.]{
\begin{tabular}{lcccc}
\multicolumn{1}{c}{\textbf{MRI Feature}} & \textbf{Original} & \textbf{Generated} & \textbf{T-Statistic} & \textbf{P-Value} \\ \hline
DWI\_Tumor   & MB & MB to MB & -0.0605 & 0.9521            \\
T1CE\_Ratio  & MB & MB to MB & -0.3643 & 0.7177            \\
T1\_Ratio    & MB & MB to MB & -0.0282 & 0.9776            \\
T1CE\_Tumor  & MB & EP to MB & -2.4975 & 0.0156            \\
FLAIR\_Ratio & MB & EP to MB & 0.4532  & 0.6524            \\
FLAIR\_Tumor & MB & EP to MB & 0.7917  & 0.4331            \\
ADC\_Ratio   & MB & PA to MB & -0.4017 & 0.6887            \\
FLAIR\_Ratio & MB & PA to MB & 11.5615 & \textless{}0.0001 \\
ADC\_Tumor   & MB & PA to MB & -4.3706 & \textless{}0.0001 \\
FLAIR\_Ratio & MB & BG to MB & 6.7026  & \textless{}0.0001 \\
ADC\_Tumor   & MB & BG to MB & -5.2062 & \textless{}0.0001 \\
ADC\_Ratio   & MB & BG to MB & -2.7487 & 0.0071           
\end{tabular}}}
\quad
\resizebox{0.482\textwidth}{!}{%
\subfloat[Difference between original EP and generated EPs.]{%
\label{EP}
\begin{tabular}{lcccc}
\multicolumn{1}{c}{\textbf{MRI Feature}} & \textbf{Original} & \textbf{Generated} & \textbf{T-Statistic} & \textbf{P-Value} \\ \hline
FLAIR\_Tumor & EP & MB to EP & -3.2397 & 0.0061            \\
ADC\_Tumor   & EP & MB to EP & -2.0273 & 0.0495            \\
ADC\_Ratio   & EP & MB to EP & -0.6434 & 0.5266            \\
FLAIR\_Tumor & EP & EP to EP & -1.0603 & 0.3018            \\
ADC\_Tumor   & EP & EP to EP & -1.4653 & 0.1519            \\
DWI\_Ratio   & EP & EP to EP & -0.4937 & 0.6278            \\
T2\_Ratio    & EP & PA to EP & 2.956   & 0.007             \\
T2\_Tumor    & EP & PA to EP & 0.3621  & 0.72              \\
T1CE\_Tumor  & EP & PA to EP & -1.1672 & 0.262             \\
T1CE\_Ratio  & EP & BG to EP & -2.1967 & 0.0428            \\
T1\_Tumor    & EP & BG to EP & -5.9549 & \textless{}0.0001 \\
DWI\_Ratio   & EP & BG to EP & 0.0059  & 0.9954           
\end{tabular}}}%
\quad
\resizebox{0.482\textwidth}{!}{%
\subfloat[Difference between original PA and generated PAs.]{%
\label{PA}
\begin{tabular}{lcccc}
\multicolumn{1}{c}{\textbf{MRI Feature}} & \textbf{Original} & \textbf{Generated} & \textbf{T-Statistic} & \textbf{P-Value} \\ \hline
T2\_Ratio    & PA & MB to PA & -2.0667 & 0.0430            \\
T2\_Tumor    & PA & MB to PA & -0.1256 & 0.9004            \\
ADC\_Tumor   & PA & MB to PA & 4.4019  & \textless{}0.0001 \\
T2\_Tumor    & PA & EP to PA & -1.3925 & 0.1694            \\
T2\_Ratio    & PA & EP to PA & 1.6279  & 0.1091            \\
ADC\_Tumor   & PA & EP to PA & 3.6692  & 0.0005            \\
ADC\_Tumor   & PA & PA to PA & 0.2781  & 0.7822            \\
T1\_Ratio    & PA & PA to PA & -0.8599 & 0.3948            \\
FLAIR\_Tumor & PA & PA to PA & -1.2497 & 0.2176            \\
T1CE\_Ratio  & PA & BG to PA & 1.711   & 0.0944            \\
T1CE\_Tumor  & PA & BG to PA & 1.7524  & 0.0862            \\
T2\_Ratio    & PA & BG to PA & 2.2029  & 0.0326           
\end{tabular}}}%
\quad
\resizebox{0.479\textwidth}{!}{%
\subfloat[Difference between original BG and generated BGs.]{%
\label{BG}
\begin{tabular}{lcccc}
\multicolumn{1}{c}{\textbf{MRI Feature}} & \textbf{Original} & \textbf{Generated} & \textbf{T-Statistic} & \textbf{P-Value} \\ \hline
T2\_Tumor   & BG & MB to BG & -2.4723 & 0.0149            \\
ADC\_Tumor  & BG & MB to BG & 5.6539  & \textless{}0.0001 \\
T1CE\_Ratio & BG & MB to BG & -6.6115 & \textless{}0.0001 \\
ADC\_Tumor  & BG & EP to BG & 2.8215  & 0.0066            \\
DWI\_Ratio  & BG & EP to BG & 0.806   & 0.4236            \\
T1CE\_Ratio & BG & EP to BG & -4.4199 & \textless{}0.0001 \\
T2\_Ratio   & BG & PA to BG & 6.78    & \textless{}0.0001 \\
T1CE\_Tumor & BG & PA to BG & -5.3162 & \textless{}0.0001 \\
T1CE\_Ratio & BG & PA to BG & -6.9185 & \textless{}0.0001 \\
ADC\_Tumor  & BG & BG to BG & -0.3252 & 0.7467            \\
DWI\_Tumor  & BG & BG to BG & -0.8181 & 0.4176            \\
DWI\_Ratio  & BG & BG to BG & -0.7461 & 0.4599           
\end{tabular}}}
\vspace{0.2cm}
\caption{The results of hypothesis tests comparing the original data with the generated data.} \label{StatsTestTable}
\end{table}

\begin{figure}[htb!]
\begin{center}
\includegraphics[width=\linewidth]{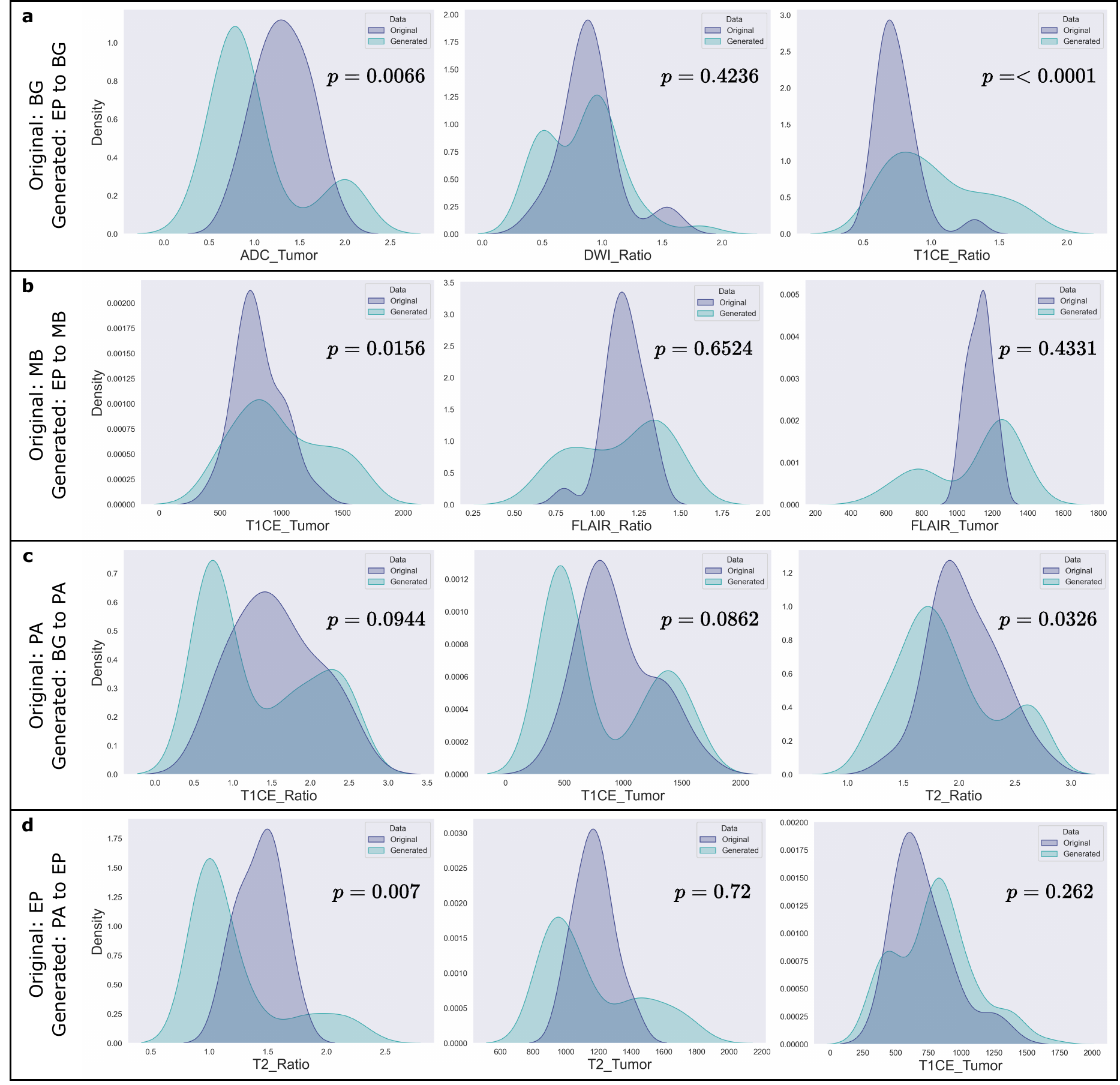}
\caption{The distributions of the original data and the generated data.} \label{Fig:Fig4}
\end{center}
\end{figure}

To validate the statistical fidelity of the counterfactual tumors generated from the original tumors, we performed a dependent T-test to evaluate the statistical difference between the tumor instances as explained in Section \ref{statisticalAn}. Although the generated counterfactuals are expected to have similar statistical properties to samples from the same class due to their minimal distance from the original instances, it is important to verify this with statistical analysis, as small borderline feature changes can lead to different classes.
The statistical similarity obtained when each tumor is transformed to represent the "what if?" scenario of other tumors. More specifically, when we transform tumor $x$ to tumor $x'$=$y$, we know that $x'$ is still dependent on x. Therefore, we measure how similar $x'$ is to the original distribution of $y$ on the feature where $x$ undergoes the most significant change. The results of statistical hypothesis tests are presented in Table \ref{StatsTestTable}. 

A high $p$-value indicates that we do not reject the difference, implying that the counterfactual explanations we generate sufficiently resemble the original distribution for that particular feature. 
In other words, for many features the difference between the generated sample and the actual data distribution is statistically insignificant. This is a desirable property for data augmentation as it means that the augmented data samples resemble similar statistical distributions.

Apart from the PA to MB transition (e.g., $p$=0.04763, $p$=0.0307), no significant differences were observed in other tumor transitions. This result can be attributed to both the fundamental optimization principle of minimizing changes during counterfactual generation and the distribution distances shown in Fig. \ref{Fig:Fig3}. Specifically, Fig. \ref{Fig:Fig3}a demonstrates a distinct separation in the distributions during the PA to MB transition, requiring a significantly larger change for transformation.

As expected, when attempting self-transformation on each tumor type, the obtained $p$-values were notably high. Evaluating at a significance level of 0.05, several features closely aligned with the actual feature distribution of the patients, making them indistinguishable from the ground truth. The following features exhibited this characteristic: FLAIR\_Ratio and FLAIR\_Tumor in the case of transforming EP to MB, ADC\_Ratio when transforming PA to MB, ADC\_Ratio during the transformation from MB to EP, T2\_Tumor and T1CE\_Tumor in the context of PA to EP transformation, DWI\_Ratio when transforming BG to EP, T2\_Tumor for MB to PA transformation, T2\_Tumor and T2\_Ratio in the case of EP to PA transformation, T1CE\_Ratio and T1CE\_Tumor during BG to PA transformation, and DWI\_Ratio when transforming EP to BG. Fig. \ref{Fig:Fig4} presents some of these cases along with their KDE distributions.

\subsection{Pushing the Boundaries of Data Augmentation through Alternative Realities}
During the construction of counterfactuals, we employed downsampling for MB and BG to align with the number of PA patients (25) during training, considering it appropriate. EP had a count of 11, and we did not increase it. The baseline results for this scenario can be observed in Table \ref{genTable}. For evaluation, the train-test splitting was conducted with a ratio of 45\% for the baseline dataset, 35\% for EP augmentation, and 25\% for EP-PA-BG augmentation.
\renewcommand{\arraystretch}{1.3}
\begin{table}[htb!]
\centering
\begin{tabular}{cccccc}
        &  \textbf{Training set} & \textbf{Test set} & \textbf{Precision} & \textbf{Recall} & \textbf{F1 Score}    \\ \hline
Baseline & 47(Real only) &  39(Real only) & 73.15 ± 9.48       & 72.20 ± 4.78    & 71.28 ± 5.62    \\
A        & 65(56 R, 9 CF) &  35(30 R, 5 CF) & 84.83 ± 4.95       & 83.75 ± 3.72    & 83.34 ± 3.65      \\
B       & 126(84 R, 42 CF) &  42(28 R, 14 CF) & 86.31 ± 4.57       & 84.64 ± 4.69    & 84.85 ± 4.72      \\
C       & 124(68 R, 56 CF) &  44(Real only) & 73.58              & 72.73           & 72.04            
\end{tabular}\vspace{0.2cm}
\caption{The impact of data augmentation using counterfactuals on classification scores is presented in the table. In the table, CF stands for counterfactuals whereas R is abbreviation for real samples. \textbf{(A)} For the first augmentation scenario, only EP counterfactuals are added, resulting in a dataset with 25 samples each for MB, EP, PA, and BG. \textbf{(B)} In the second augmentation scenario, counterfactuals for EP, PA, and BG are added to balance the number of samples with original count of MB. Assuming all counterfactual examples represent real data, this scenario results from a dataset comprising 42 samples each for MB, EP, PA, and BG. \textbf{(C)} The third scenario involves moving all real samples to the test set, with 11 patients in each category. Consequently, no factual EP samples are left in the training set, and the model is trained accordingly. } \label{genTable}
\end{table}

To address the data imbalance, we examined the inclusion of generated counterfactuals for data augmentation. For example, by equalizing EP with the other tumor types and incorporating 14 different generated counterfactuals alongside the originals, we excluded EP-to-EP instances. Opting for transitions from various tumor types to maximize variance and generalizability, we achieved an improvement of up to 12.06\% as shown in Table \ref{genTable}, case A.

To incorporate the previously set aside MB and BG data, we aligned all tumor types, except themselves, with counterfactuals generated from different tumor types. BG, PA, and EP were included with MB, and all were evaluated as a group of 42 patients, which was the maximum patient count for one tumor type. When considering the counterfactuals as actual patients, the outcomes align with the results presented in Table \ref{genTable}, case B.

Furthermore, in the case examined in Table \ref{genTable}, case C, 11 patients were included from each tumor type in the test set, resulting in no actual EP patients in the training set. Consequently, during training, we had 31 real samples for MB, 0 real and 31 counterfactual samples for EP, 14 real and 17 counterfactual samples for PA, and 23 real and 8 counterfactual samples for BG. Notably, when evaluating on real samples, the results were intriguing. Despite the absence of real EP patients in the training data, the model successfully identified 5 out of the 11 patients, leading to an overall baseline score that was, on average, 0.76\% higher. 

In all cases, LR consistently yielded the best performance in terms of prediction accuracy and execution time. The boosting algorithms took significantly more time to converge on the predictions, although they did not improve the prediction performance. Therefore, to simplify the presentation of the results, we only reported the results of the LR classifier in the table.

\section{Discussion}

Spatial heterogeneity in pediatric brain tumors, especially from the posterior fossa, complicates accurate differentiation \cite{15porto2014conventional, 16koob2014cerebral, 17orphanidou2014texture, 18moharamzad2018brainstem, 19d2018differential, 20duc2019magnetic, 21duc2020role, reddy2020pediatric, chen2023apparent}. Accurate diagnoses are essential since each tumor requires specific treatments impacting patient outcomes. While AI advancements in medical imaging are promising, their black-box nature hinders clinical adoption. Our study introduces a novel approach, leveraging counterfactual explanations to interpret MRI features, aiming to provide clinicians an intuitive tool. This research pioneers feature-based counterfactual investigations in pediatric brain tumors.

The medical literature highlights that individualized care is crucial, aligning with personalized healthcare \cite{chawla2013bringing, shaban2018health, paranjape2020short, johnson2021precision}. The ability to generate hypothetical scenarios for a patient based on MRI features offers a significant advantage in medical diagnosis. By creating what-if scenarios, radiologists are equipped with additional intuitive data, enhancing their decision-making ability. In the proposed approach, counterfactual explanations can be seamlessly integrated into the radiologist's existing clinical workflow as a decision support system and provide support to the personalized treatment process for patients. This approach could potentially mitigate the need for invasive procedures and provide a clearer perspective on the tumor's nature based on MRI data alone. Our method produces tailored explanations for each patient, drawing from past cases, facilitating understanding of tumor differentiation based on MRI data.

The concept of counterfactuals, which has been debated in philosophy and psychology for decades, has found its place in the field of artificial intelligence under various names. Though the idea has historical roots, its comprehensive implementation in AI is a more recent phenomenon. Like many preceding studies, we have adapted this concept for the clinical domain, providing valuable insights for clinicians. Furthermore, our work enriches the literature on medical counterfactuals by offering a unique perspective tailored to specific tasks. Through counterfactuals, we demonstrate alternative possibilities within the decision space and elucidate the rationales behind specific decisions pertaining to pediatric patients.

To the best of our knowledge, no prior studies have addressed counterfactuals regarding posterior fossa tumors. We filled this gap and subjected results to statistical tests, presented in Section \ref{section4-3}. We explored the utility of counterfactuals both as post-classifiers and indicators of significant MRI features. The LR model was the most effective, hence we used it for counterfactual generation.

We developed a framework aimed at enhancing the utility of counterfactuals beyond what DiCE offers for our case. When faced with a sizable patient pool, utilizing another counterfactual algorithm considering a subsample and substituting excluded patients can aid in statistical testing. In certain cases, the alternate methods could optimize within a more favorable timeframe than DiCE.

Our approach, which utilizes counterfactual explanations in a classifier-like manner, eliminates the need to separate different test sets. Consequently, the performance of our machine learning models significantly exceeds the baseline scores, with only a few patients excluded from the decision space to simulate the scenario of newly arriving patients. In this scenario, all training samples serve as test patients as we explore the decision space. To accomplish this, DiCE provides valuable information about misclassified samples, allowing us to exclude the associated counterfactuals from the statistical analysis through post-processing. 

Fig. \ref{Fig:fig2} and Table \ref{table2} depict a hypothetical scenario involving a patient with an initially unknown EP tumor. The radiologist examining the MR images was uncertain about whether the tumor was of the MB or EP type. A key challenge in such cases is the lack of additional information, which often necessitates invasive procedures like brain surgery and tissue sampling for histopathological analysis to obtain a definitive diagnosis. To overcome this issue, we generate alternative scenarios based solely on the MRI features. These scenarios provide additional quantitative information to the radiologist, enabling them to assess the response based on the individual's biological characteristics.

Moreover, Table \ref{table4} demonstrates the efficacy of our approach in identifying patients with diverse tumor types that were previously unidentified and not encompassed within the decision space. While ML models can also accomplish this task, our method offers an additional advantage by preserving information regarding tissue characteristics, which in turn reveal similarities or differences among tumors. Additionally, our approach calculates distances to other tumors by transforming the features into a uniform distribution through standard scaling, providing valuable insights about the proximity. This valuable information aids in our comprehension of the differentiation among tumors in the dataset.

Table \ref{table5} presents the total count of modifications made to susceptible features, with the exception of the parenchymas that serves as reference points, when generating samples for different patients. The statistical report enables a human verification of the optimization process, wherein minimal changes are implemented to achieve the desired outcome. It also confirms that the features exhibiting the highest variations during the generation of alternative realities are those with the most distinct distributions between two tumors. To elucidate the analysis of their distributions, we present Fig. \ref{Fig:Fig3} as a visual representation. Table \ref{best3} presents the top three most variable features extracted from the reports obtained for all tumor matches in Table \ref{table5}. This provides a more concise overview of all the cases.

Table \ref{StatsTestTable} exhibits a statistical analysis demonstrating the high degree of similarity between the generated data and reality across different data spaces, specifically focusing on the most frequently selected features. A high $p$-value indicates that the generated samples cannot be well distinguished, implying the effectiveness of the independent transformation process, which produces significant alternative realities separate from the original space. Fig. \ref{Fig:Fig4} illustrates an example of some transformations from Table \ref{StatsTestTable}, displaying their corresponding $p$-values, as well as the kernel density estimation of the generated data in comparison to the original data.

Our generated counterfactuals offer potential advantages for data augmentation. Traditional methods, like SMOTE \cite{chawla2002smote}, fall short in real-world alignment and interpretability. We suggest counterfactuals as a viable alternative, particularly when data is limited. As shown in Fig. \ref{genTable}, while case C cannot be benchmarked directly due to added test patients, the inclusion of more real samples enhances outcomes. This improvement aligns with findings in \cite{tanyel4421493deciphering}. However, challenges arise when certain EP patients, which complicate differentiation, are considered. Despite these challenges, the ability to make accurate predictions for many patients without actual EP training data highlights the promise of our approach and suggests directions for future research.

Counterfactual explanations can also address model bias in medical diagnoses \cite{mikolajczyk2021towards, wang2023towards}. Ensuring fairness and transparency in decision-making processes is vital, suggesting the value of counterfactuals in this direction.

\subsection{Potential Challenges and Limitations}


One of the limitations encountered in this study is the size of MRI data collected. Considering the low incidence of this disease, it is difficult to encounter a large number of patients for each type of pediatric posterior fossa tumor in a single hospital. Although we observe that the current data set provides valid counterfactual explanations and clinically useful results, there is still a need for large research data sets collected at national or international level in this field. 

Expanding the dataset's scope and size may lead to the potential for counterfactual explanations to emerge with new and different latent features. Furthermore, the use of a comprehensive data set that captures a broader range of scenarios for pediatric posterior fossa tumors encountered in clinical practice may play an important role in ensuring a wider applicability and confirming the generalizability of the counterfactual explanations currently uncovered.


There are recognized challenges in applying the DiCE method to diverse datasets, sometimes resulting in extended optimization times and difficulties achieving convergence. Addressing these challenges will be an essential step forward. Exploring alternative methodologies and delving deeper into the vast landscape of counterfactual algorithms might also be beneficial. 

To further advance the field, future research should consider incorporating additional advanced MRI protocols, enriching our understanding and diagnostic capabilities regarding pediatric posterior fossa tumors.

\section{Conclusion}
This paper presents a novel interpretability approach in medical research, using pediatric posterior fossa brain tumors as a case study. By generating counterfactual explanations, it delivers personalized insights, validates predicted outcomes, and highlights how predictions vary under different conditions. Although medical regulations and workflow concerns remain hurdles, the use of explainable AI in medicine is poised to grow as its benefits become clearer.

Our method bridges the gap between machine learning and clinical decision-making, potentially leading to better patient outcomes. Further research is needed to integrate counterfactual explanations into clinical practice and evaluate their real-world performance. Larger studies, including different diseases, could produce even more robust “alternative realities” from MRI features. Overall, we believe this approach has the potential to shift the perspective of radiologists and other medical professionals by offering more human-like, actionable insights.

\begin{credits}
\subsubsection{\ackname} The authors received no financial support for the research, authorship, and/or publication of this article.

\subsubsection*{\discintname} The authors declare that they have no known competing financial interests or personal relationships that could have appeared to influence the work reported in this paper.

\subsection*{Institutional Review Board Statement} After obtaining approval from the Institutional Review Board of Children Hospital of 02 with approval number [Ref: 632 QÐ-NÐ2 dated 12 May 2019], we conducted the study in both Radiology and Neurosurgery departments in accordance with the 1964 Helsinki declaration.

\subsection*{Data \& Code Availability}
The datasets generated and/or analyzed during the current study are not publicly available due to privacy concerns but are available from the Dr. Keserci upon reasonable request. The source codes of the presented study can be accessed at:\\ \href{https://github.com/tanyelai/counterfactual-explanations-for-medical-research}{https://github.com/tanyelai/counterfactual-explanations-for-medical-research}

\end{credits}
%
%
%
\bibliographystyle{splncs04}
\bibliography{paper-0034}

\end{document}